\documentclass[runningheads]{llncs}
\usepackage[T1]{fontenc}
\usepackage[dvipsnames]{xcolor}
\usepackage{tikz}
\usepackage{tikzscale}
\usepackage{graphicx}
\usepackage{adjustbox}
\usepackage{orcidlink}
\usepackage[normalem]{ulem}
\usepackage{hyperref}
\usepackage{booktabs}
\usetikzlibrary{arrows.meta, positioning, shapes, backgrounds, patterns, calc, decorations.pathreplacing, decorations.pathmorphing}

\begin{document}
\title{Benchmarking that Matters: Rethinking Benchmarking for Practical Impact}
\author{
Anna V. Kononova\inst{1}\orcidlink{000-0002-4138-7024} \and 
Niki van Stein\inst{1}\orcidlink{0000-0002-0013-7969} \and 
Olaf Mersmann\inst{2}\orcidlink{0000-0002-7720-4939} \and
Thomas Bäck\inst{1}\orcidlink{0000-0001-6768-1478} \and 
Thomas Bartz-Beielstein\inst{3}\orcidlink{000-0002-5938-5158} \and
Tobias Glasmachers\inst{4}\orcidlink{0000-0003-1886-1} \and 
Michael Hellwig\inst{5}\orcidlink{0000-0002-6731-8166} \and 
Sebastian Krey\inst{6} \and
Jakub K\r{u}dela\inst{7}\orcidlink{0000-0002-4372-2105} \and 
Boris Naujoks\inst{3}\orcidlink{0000-0002-8969-4795} \and 
Leonard Papenmeier\inst{8}\orcidlink{0000-0001-9338-1567} \and 
Elena Raponi\inst{1}\orcidlink{0000-0001-6841-7409} \and 
Quentin Renau\inst{9}\orcidlink{0000-0002-2487-981X} \and 
Jeroen Rook\inst{10}\orcidlink{0000-0002-3921-0107} \and 
Lennart Schäpermeier\inst{8}\orcidlink{0000-0003-3929-7465} \and 
Diederick Vermetten\inst{11}\orcidlink{0000-0003-3040-7162} \and 
Daniela Zaharie\inst{12}\orcidlink{0000-0003-3388-6058} 
}
\authorrunning{
A. V. Kononova et al.
}
\institute{
LIACS, Leiden University, the Netherlands \email{a.kononova@liacs.leidenuniv.nl}\and 
Hochschule des Bundes für öffentliche Verwaltung, Germany\and 
TH Köln, Germany\and 
Ruhr-Universität Bochum, Germany\and 
Vorarlberg University of Applied Sciences, Austria\and 
GWDG, Göttingen, Germany\and
Brno University of Technology, Czech Republic\and 
University of Münster, Germany\and 
University of Stirling, UK\and 
Paderborn University, Germany\and 
Sorbonne Université, CNRS, LIP6, France\and 
West University of Timișoara, Romania 
}
\maketitle              
\begin{abstract}
Benchmarking has driven scientific progress in Evolutionary Computation, yet current practices fall short of real-world needs. Widely used synthetic suites such as BBOB and CEC isolate algorithmic phenomena but poorly reflect the structure, constraints, and information limitations of continuous and mixed-integer optimization problems in practice. This disconnect leads to the misuse of benchmarking suites for competitions, automated algorithm selection, and industrial decision-making, despite these suites being designed for different purposes.

We identify key gaps in current benchmarking practices and tooling, including limited availability of real-world-inspired problems, missing high-level features, and challenges in multi-objective and noisy settings. We propose a vision centered on curated real-world-inspired benchmarks, practitioner-accessible feature spaces and community-maintained performance databases. Real progress requires coordinated effort: A living benchmarking ecosystem that evolves with real-world insights and supports both scientific understanding and industrial use.

\end{abstract}

\section{Introduction}\label{sec:sota}
Benchmarking has been intertwined with Evolutionary Computation since the very beginning of the field, more than sixty years ago~\cite{JFS97,back2023evolutionary}. As stochastic optimizers matured, it became evident that meaningful conclusions about algorithm performance require systematic, repeated experimentation. Over the years, this motivated the creation of widely used benchmark suites, most prominently BBOB\footnote{\url{https://coco-platform.org/testsuites/bbob/overview.html}}~\cite{hansen2021coco,bbob2019}, CEC~\cite{vskvorc2019cec}, the IOH suite\footnote{\url{https://iohprofiler.github.io/IOHanalyzer/}}~\cite{IOHprofiler}, which advanced empirical methodology for continuous black-box optimization. Yet, despite this long history, benchmarking in continuous and mixed-integer optimization is still far from ideal. Compared to the discrete optimization domain, where benchmarking culture is well established and diverse real-world inspired (RWI) problem collections exist, the continuous domain remains underserved: benchmarking practices are uneven, the available test suites cover only a narrow subset of structural problem characteristics, and their connection to industrial needs is weak.

A central reason is that benchmarking serves fundamentally different purposes in academia and industry. Academic benchmarking is oriented towards \emph{knowledge generation}: understanding why algorithms behave as they do, comparing solvers under controlled variations and validating theoretical insights~\cite{bart20gArxiv}. In contrast, industrial benchmarking functions as a \emph{decision-support process}~\cite{chas10a}: the goal is not general insight but selecting a reliable solver for a single, costly problem instance, often under tight evaluation budgets and with incomplete problem information.

These diverging objectives expose limitations in today’s benchmarking landscape. Most widely used academic suites are \emph{synthetic} and deliberately constructed to isolate algorithmic phenomena: functions are chosen to highlight separability, ill-conditioning, multimodality, or plateaus and not to resemble real-world optimization landscapes~\cite{rardin2001experimental,johnson2002experimental}. As a result, they provide deep insights into algorithmic behaviour but offer little guidance for practitioners deciding which solver will perform well on a specific engineering or simulation-based problem.

At the same time, RWI benchmarks remain fragmented and scarce. Existing collections tend to focus on individual engineering domains and often feature fixed dimensionality, strong constraints or expensive simulations. Important high-level problem characteristics such as multimodality, continuity, variable interactions or meaningful parameter semantics, are frequently unknown or not systematically represented. This gap hampers reproducibility, interpretability and the ability to match solver behaviour to practical problem classes.

Finally, the tooling ecosystem has grown around the needs of expert algorithm designers and not industrial users. Current benchmarking platforms provide strong support for performance visualization but limited guidance on research–driven interpretation, limited validation of experimental data and little support for curating collections of RWI problems. This misalignment contributes to the misuse of synthetic suites, for competitions, solver recommendation and industrial decision-making, despite their original design goals.

\paragraph{Scope of this paper.}
This paper focuses exclusively on benchmarking for \textbf{continuous and mixed-integer} black-box optimization. We explicitly do not address combinatorial optimization, whose benchmarking traditions and structural challenges differ substantially. We aim to (i) articulate the main gaps preventing current benchmarking practices from achieving real-world relevance, (ii) outline a principled vision for designing, curating, and deploying real-world inspired benchmarks that support both scientific progress and industrial decision-making.

\section{Purposes of Benchmarking}\label{sec:purposes}
Benchmarking is the practice of empirically evaluating the performance of computational algorithms. 
Its importance for optimization algorithms is cemented by the ``No Free Lunch'' theorems~\cite{NFLT,Rowe2009}, which proved that no universally superior optimization algorithm exists.
This finding invalidates the quest for a single ``best'' method, establishing that an algorithm's performance is scientifically meaningless without the context of the specific problems it is applied to.
However, the term ``benchmarking'' has \textit{different} meanings for different communities.
It describes at least two distinct paradigms, each driven by different goals, constraints and metrics (see Section~\ref{sec:vision} for more detail):
\begin{itemize}
    \item academic world of scientific research, with the primary objective of \emph{knowledge generation},
    \item industrial world of real-world application, with the primary objective of \emph{actionable decision support} to solve a unique problem~\cite{chas10a}.  
\end{itemize}

The academic paradigm is best understood through the taxonomy of scientific goals as codified in the \emph{Benchmarking in Optimization: Best Practice and Open Issues} paper~\cite{bart20gArxiv}, organizing benchmarking activities into five primary objectives, namely: visualization and basic assessment, sensitivity of performance, performance extrapolation, theory, and algorithm development.

\begin{table}[!t]
\centering
\caption{Comparison of scientific (academic), as introduced in~\cite{bart20gArxiv}, and industrial benchmarking goals.}\label{tab:goals-comparison}
\small
\begin{tabular}{lp{0.39\textwidth}p{0.43\textwidth}} \toprule
\textbf{Goal} & \textbf{Academia} & \textbf{Industry} \\ \midrule
Assessment & How well does the algorithm perform across problem sets? & De-risk investment: will it work? (Technology Due Diligence)\\ \midrule
Sensitivity & Why does it perform well? Analyze parameters and features & Does it work for \emph{my} specific problem? (Application-Specific Validation)\\ \midrule
Extrapolation & Train models for automated algorithm selection/configuration & Forecast ROI and business value (Building the Business Case)\\ \midrule
Theory & Validate or inspire theoretical analysis. & Not relevant (focus on outcomes) \\ \midrule
Development & Iterative tool for algorithm refinement and validation & Reduce costs and time-to-market (Process Optimization)\\ \bottomrule
\end{tabular}
\end{table}

We argue that these scientific goals must be reformulated for the industrial setting, which operates under a different set of imposed constraints.
Academic studies can often afford thousands of runs on abstract test functions, but many industrial applications face \emph{very limited} time frames and \emph{prohibitively expensive} objective function evaluations.

\subsection{Good Benchmarks}
Regardless of user perspective, benchmarks play a central role in the evaluation and development of problem-solving techniques. They provide standardized, comparable and reproducible conditions for the rigorous evaluation of available algorithms~\cite{bart20gArxiv}. 
A good benchmark environment enables objective comparisons, promotes innovation, helps to systematically identify the strengths and weaknesses of different approaches and allows to extrapolate algorithm performance to application problems. It ideally contains useful information about the problem structure, e.g. the semantic meaning of variables, the origin of the problem, or relevant constraints and assumptions about the situation. This information is vital for analyzing algorithm behavior and understanding why certain methods work better than others. Benchmarking needs to reflect typical challenges in the target area and cover a wide range of similar problem types~\cite{rardin2001experimental}. Such problem collections are usually expected to vary in terms of difficulty and structure. The respective tools need to be structured in such a way that they directly support the transparency and integrity of research activities and prevents false conclusions and bias~\cite{johnson2002experimental}.
Although many of these requirements for thorough benchmarking are generally valid, some aspects differ slightly between academic and industrial contexts.

\paragraph{Research Perspective. }
From an analytical perspective, it is desirable to create benchmarking environments that reflect the diversity of a specific theoretical domain. Such problems must be systematically designed on the basis of clear mathematical properties in order to gain a precise understanding of how algorithmic operators work and to explore the potential for generalizing solvers~\cite{johnson2002experimental,whitley1996evaluating}. The problem collection needs to be structured in a way that allows for ablation studies by allowing for comparisons between pairs of functions that differ in specific, well-defined properties of an established taxonomy. To ensure clarity and consistency, the taxonomy needs to be thoroughly explained and documented to avoid ambiguity in interpretation. The benchmarks are expected to be linked to known algorithmic (and theoretical) results to enable meaningful comparisons and validations of new methods. Ideally, they provide known optimal values and the position of the optimizer in order to precisely evaluate the absolute quality of the solutions. Moreover, the problem dimensionality should be scalable to test the performance of algorithms for different problem sizes. To enable extensive experimentation, the evaluations are also usually demanded to be rather computationally inexpensive.

To account for algorithmic bias and the stochasticity of meta-heuristics, each problem should be represented by an adequate number of corresponding instances. In this context, a problem instance is defined as a slight variation of the basic problem, i.e. through translation, rotation, and scaling. 
An algorithm can therefore be executed on multiple instances of a problem. Performance across different instances of the same function can be assumed to be stable. This design prevents individual operators from exploiting fixed optimal positions or axis directions, thereby promoting robustness and generalization in performance evaluation. An established scheme for such an instantiation is provided by the well-known BBOB test suite~\cite{bbob2019}, which provides a precise framework for the comparison and analysis of different algorithmic working principles.

\paragraph{Practitioner Perspective. }\label{sec:practitioner}
For practice-oriented users, the requirements for useful benchmarking environments are shifted more towards achieving improved solution quality in a specific problem setting~\cite{DBLP:conf/gecco/MersmannBTPWR11}. In contrast to systematic, mathematically motivated problem gradations, the focus is often on very particular, static problems, which are based on realistic use cases such as real user data, production data or simulated scenarios from practice. 
To reflect this, benchmarks need to be aligned as closely as possible with real-world requirements and constraints like resource and runtime limits, noise or incomplete information.

As RWI problems often have no known optima, best-known reference solutions are highly appreciated, yet difficult to determine and therefore less common~\cite{beiranvand2017best}. One reason for this shortage is that evaluations of real-world objective functions are generally much more costly due to operating/simulation efforts. 
Benchmarking should therefore ideally be designed in such a way that functional evaluations can be carried out as fast and effortlessly as possible, but without losing the characteristics of the real problem. 

The benchmarking set is also required to include a comprehensive understanding of the problem characteristics and a detailed explanation of the importance of individual search space parameters for each RWI test function~\cite{HELLWIG2019bench}. Since the creation of problem instances is anything but straightforward in the face of real-world problems (e.g. due to complicated or lacking analytical descriptions), falling back to problem clusters of somewhat similar objective function landscapes might offer a pragmatic solution. 
Users would then be able to screen the performance results on those RWI benchmark problems that most closely resemble their actual real-world problem and select the most successful solver for their purpose.
Additionally, benchmarking for practical use should enable the assignment of noise to multiple sources of measurement errors and, ideally, the quantification of uncertainty, since benchmarks are noisy due to the inherent variability in algorithm performance, stochastic problem elements and environmental factors affecting reproducibility and reliability.

Although the requirements for convenient RWI benchmarks can be clearly formulated, there are hardly any usable problem collections, let alone systematic benchmarking environments.

\subsection{Current Usage of RWI Problems in Benchmarking}
Synthetic optimization problems have been intensively used to benchmark the behavior of optimization methods and to obtain insights into their performance. However, these insights are not always transferable to real-world problems because the characteristics of synthetic problems, which are usually explicit, do not necessarily match the characteristics of real-world problems, which are not easy to observe.  
Traditional RWI test suites are merely collections of problems corresponding to the same or related application domains. Most problems included in such test suites, e.g. \cite{He2020,KUMAR2020100693,Tanabe2020}, are related to engineering (e.g. bar truss, beam, vessel or car cab design; optimization of power systems etc.) and are frequently characterized by bounded domains for the design variables, nonlinear constraints, and specific (e.g. physically imposed) values of the parameters. These can potentially create difficulties in controlling the problem complexity level and generating problem instances. There are, however, classes of problems for which realistic and scalable instances that cover different characteristics of the fitness landscape have been proposed (e.g. game optimization \cite{Volz2019}, multi-objective design of actuators \cite{Picard2021}). 

Moreover, the whole frameworks have been recently designed that aim to generate test instances inspired by real-world problems. For instance, \cite{Thomaser2023} proposes a strategy to generate an arbitrary number of different synthetic functions that are cheap in evaluation and mimic the characteristics of several 2D vehicle dynamics problems. Another framework, aiming to generate instances of a generic camera optimization problem, is {\it Ealain} (\cite{Renau2024}). It allows the generation, by changing the parameters of the environment corresponding to the problem, of instances belonging to different classes of problems (single-objective, multi-objective, multi-fidelity, constrained).
Even if these generators are based on different synthesis strategies, they focus on flexibility and on ensuring a good resemblance with the real-world problem. In the case of the offline black-box optimization, there are some recent benchmarks (Design-Bench \cite{Trabucco2022}, SOO-Bench \cite{Qian2025}) that include RWI problems from scientific and engineering disciplines. 
 
A recent repository\footnote{https://openoptimizationorg.github.io/OPL/} collecting information about existing test problems, suites of problems, and problem generators, currently contains 63 entries, out of which only 15 correspond to RWI problems. The current status of RWI benchmarking suggests the need for a more principled approach in designing benchmarking environments for practice-oriented users.

\section{Misuse}\label{sec:misuse_bbob}
Benchmarking practices have evolved significantly over the past few decades, driven by the increasing availability of diverse problem suites. While this progress has enabled more robust algorithm evaluation, the widespread adoption of certain suites -- most notably COCO’s BBOB~\cite{hansen2021coco} -- can sometimes obscure the original objectives envisioned by their designers.
In the following sections, we discuss two of the largest challenges in benchmarking black-box optimizers.

\paragraph{Usage of Synthetic Benchmark Suites for Competition.}
Generally, academic benchmark problems are designed from the perspective of algorithm design. Functions are included in a suite because the performance achieved by an algorithm reveals some useful insights into that algorithm's behavior in a controlled environment.

For a real-world driven benchmark study, however, the overall purpose of the experiment is generally \textit{different}.
Instead of understanding the behavior of an algorithm, the focus is rather on finding a single `best' method that performs well on a broad set of problems representative of those one expects in the `real world'. This leads to a scenario where aggregation is overused to obtain a clear number that distinguishes algorithms from each other, with one being the eventual `winner'. This competitive approach to benchmarking, when applied to problem suites not explicitly designed for this purpose, leads to \textit{biases} in favor of the properties of that suite. For example, from an algorithm selection perspective, the implicit assumption that train (benchmark) and test (industrial) problems come from the same underlying distribution is unlikely to hold.

One example is the popular BBOB benchmark suite~\cite{hansen2021coco}, which has been carefully designed to answer specific research questions~\cite{bbob2019}. Functions are selected to highlight specific optimization challenges and corresponding algorithm properties. For example, including a sphere function is not particularly interesting in itself, but contrasting the performance on a sphere and an ellipsoid function provides valuable information on how an algorithm is affected by a problem being ill-conditioned. For that purpose, the ellipsoid should have a very high condition number, no matter whether the number is believed to be realistic in practice or not.

However, performance data are typically presented as simple lists, e.g. with one table row or plot per benchmark problem. That style of presenting results is simple, easy to communicate and complete. The downside is that it fosters a \textit{problematic} interpretation of the results. A flat list suggests that all functions are equally informative and can be aggregated safely.
Clearly, aggregating performance data over a problem suite composed in the above way is rather meaningless, because the problems were \emph{not} selected to be representative of a certain class of problems, but each problem studies a certain aspect of a method. Therefore, the presentation of results should discourage such problematic use of the results. Quite in contrast, it should instead encourage an interpretation of the results for which the suite was designed.

\paragraph{Usage of Synthetic Benchmark Suites for Algorithm Selection.}
Algorithm selection is another example of problem suites being used for purposes for which they were \textit{not} designed.
The goal of automated algorithm selection is to find the most suitable algorithm from a portfolio to solve a problem at hand, with the underlying hypothesis that problems, described by problem features, that are close in the problem space will exhibit similar performances in the performance space.

In recent works, a large number of features have been introduced to describe optimization problems, such as~\cite{DBLP:conf/gecco/MersmannBTPWR11,DBLP:journals/swevo/PetelinCE24,DBLP:journals/corr/abs-2401-01192}.
The performance of these features, i.e. how accurate selectors built using these features, is often assessed using synthetic benchmark suites such as BBOB~\cite{hansen2021coco} or the CEC suites~\cite{CEC2017}.
Using synthetic benchmark suites may lead to some pitfalls.
As functions in these suites have been designed to gain insights into algorithm behaviors, functions often cover a wide range of landscapes.
For this reason, it has been noted that \textit{leave-one-problem-out} cross-validation on BBOB is a particularly challenging problem~\cite{DBLP:conf/foga/DerbelLVAT19}.
The most commonly used cross-validation setting is \textit{leave-one-instance-out} cross-validation, where an instance of each considered function is left out of the training data.
As such, training and testing sets contain the same functions and differ only by a transformation (rotation, translation, and/or scaling) of these functions.
It has been pointed out that: (1) this setting may be too simple with a couple of features being able to perform an almost perfect selection, (2) features can be easily generated to take advantage of this setting~\cite{DBLP:conf/evoW/RenauDDD21,DBLP:journals/swevo/CenikjPSCE25}. 

\section{Challenges of RWI Benchmarks}
In light of these misinterpretations and the current state of real-world problem collections, there is still a long way to go before viable real-world-inspired benchmarking environments are readily established. 
Regarding the demand for good benchmarking, there are still a number of challenges that need to be addressed to accurately capture the industry perspective. This section will focus on the \textit{most significant} of these.

\subsection{Feature Information} 
Relevant problem features are a cornerstone for efficient algorithm selection. Available knowledge of the often-used high-level features, such as the number of objectives, the number and type of variables, the existence and type of constraints and other structural properties (modality, smoothness, etc.), can itself lead us to the selection of an appropriate algorithm (or a class of algorithms). For synthetic benchmark problems, such features can usually be easily derived from the problem definition. However, in real-world problems, many of these high-level features are unknown. A recent poll \cite{DBLP:series/ncs/BlomDVMNNOT23} among optimization practitioners shows that (out of the 45 problems submitted) for roughly 80\% of them, many of the important high-level features of the problems were not known (such as convexity/concavity, continuity or shape of the Pareto front). 
For 40\%, the ranges of the objective values were not known, and for 20\%, even a feasible solution was not known. For some real-world problems, there is also a question of the ``correct'' number of variables (that usually correspond to the granularity of the solution) that should be used \cite{shehadeh2025benchmarking,kuudela2024benchmarking}. For these variable dimension problems, our currently used synthetic benchmarks are inadequate.

When the high-level features are not known, at least some problem-specific numerical features might be available, such as positions of obstacles in planning problems in robotics or the load distribution for topology optimization problems. These, however, will inevitably be problem-specific and could be used for algorithm selection only within a well-defined class of problems.

The remaining option is to rely on pure data-driven feature extraction, best exemplified by the Exploratory Landscape Analysis (ELA) features \cite{DBLP:conf/gecco/MersmannBTPWR11}. These can be used to find exploitable landscape properties, such as ruggedness or plateaus in the objective function, the existence of funnel structures \cite{kerschke2015detecting} or other similar properties \cite{volz2023tools}. The computation of these features (which involves evaluation of a relatively large number of samples) might however not be possible for computationally expensive problems (in the poll referenced above, a large portion of the real-world problems had a budget of < 1000 function evaluations). Additionally, the generalizability of ELA features for algorithmic selection (across different problem classes) is not fully established \cite{lacroix2019limitations,DBLP:journals/swevo/CenikjPSCE25}.

\subsection{Challenges in Multi-objective Benchmarking}
Real-world optimisation problems are diverse and often involve characteristics such as black-box structure, mixed variable types, constraints, noise or multi-fidelity evaluations. Among these, multi-objective optimisation (MOO) provides a particularly illustrative example of the added complexity that RWI benchmarking must capture. The issues discussed below are representative of broader challenges that also arise in noisy, constrained and mixed-integer settings.

\begin{description}
    \item[Performance indicators] Since total ordering is lost in MOO, performance must be assessed using indicators that rank sets of solutions. While many indicators exist~\cite{ABC+21,LY19}, only a few are Pareto-compliant, yet this property is essential for meaningful comparison~\cite{Knowles2002_metrics}.
    \item[Archiving] MOO algorithms typically maintain an archive of non-dominated solutions, often using elitist populations~\cite{DPA+02,BNE07}. For benchmarking, it must be clarified whether comparisons are based on archives or final populations as the choice directly affects performance assessment.
    \item[Reference points] Many indicators require ideal or nadir reference points. These choices substantially influence scores~\cite{ShaEtAl21}, yet are often fixed without discussion. Clear guidelines are needed to avoid unintended biases.
    \item[Repeated runs] Unlike SO optimisation, two weak solution sets in MO can be complementary when merged, meaning that indicator-based summaries (e.g. medians) may underestimate achievable performance. This interaction is rarely accounted for in current benchmarking.
    \item[Entanglement] MOO algorithm design is tightly linked to indicators, which may be used as selection operators. When the same indicator is later used for benchmarking, evaluations can become circular or biased. Indicators should ideally not depend on algorithmic design choices such as population size.
\end{description}

Beyond these evaluation issues, current MOO test suites cover only a limited set of challenges. To support principled benchmarking a broader taxonomy of problem properties is required including:

\begin{description}
    \item[Multimodality] MOO introduces emergent properties, e.g. connected vs.\ disconnected Pareto sets or locally efficient sets, beyond those of the component SO problems~\cite{grimme2021peeking}.
    \item[Pareto front shape] The geometry of the front (convex, concave, linear) directly affects which aggregation strategies are valid, such as weighted sums only recovering extremal points for concave fronts.
    \item[Separability] Standard variation operators exploit axis alignment; robustness to rotation of the Pareto set is underexplored.
    \item[Plateaus] Flat regions of the objective space are rarely represented in existing MOO benchmarks.
    \item[Conditioning] The effect of conditioning on multi-objective landscapes remains insufficiently studied~\cite{glasmachers2019challenges}.
\end{description}

Finally, noisy MOO problems pose an additional challenge: noise affects dominance relations and stability of indicator values, yet most benchmarks assume deterministic evaluations. Accounting for this is essential for RWI benchmarking.

\subsection{Gaps in Tooling}
To facilitate RWI benchmarking, we must take a critical look at the tooling required to support this. 
While over the years the status of benchmarking support structures has improved significantly, there are still gaps to fill to provide robust tools that cover all aspects of the benchmark pipeline without imposing a single benchmarking purpose or methodology. 

Where aspects such as problem representation and logging (on the experimental side) and performance visualization and analysis (on the post-processing side) are quite well-covered, the lack of rigorous data validation and curation tools leads to a lack of trust in shared benchmarking results. 
This includes low-level aspects such as verifying that all runs have completed and no data is missing, to high-level aspects of data collection and curation with correct meta-data and their long-term storage and dissemination. 

As discussed in Section~\ref{sec:misuse_bbob}, many benchmarking frameworks were designed with a focus on an expert user base of algorithm designers, without making this assumption explicit.
This misalignment in these tools' intent and usage has been part of the growing misuse of popular benchmarks, especially when the goals are inspired by real-world use cases.
This can also be partly explained by the focus on creating large frameworks that cover the end-to-end benchmarking process in one setup, while a toolbox of interoperable components might facilitate a wider range of benchmarking purposes.  

\section{A Vision for RWI Benchmarking}\label{sec:vision}
\begin{figure}[!t]
  \centering
  \resizebox{\textwidth}{!}{%
  \begin{tikzpicture}
    \tikzset{every node/.style={font=\large}}

    \tikzset{
      labelbox/.style={
        draw, rounded corners, fill=white, align=left,
        text width=5.1cm, inner sep=5pt
      },
      narrowerlabelbox/.style={
        draw, rounded corners, fill=white, align=left,
        text width=3.8cm, inner sep=5pt
      },
      widerlabelbox/.style={
        draw, rounded corners, fill=white, align=left,
        text width=5.7cm, inner sep=5pt
      },
      nobox/.style={
        rounded corners, fill=none, align=left, text width=2.3cm, inner sep=5pt
      },
      noboxwider/.style={
        rounded corners, fill=none, align=left, text width=2.6cm, inner sep=5pt
      },
      widebox/.style={
        draw, rounded corners, fill=white, align=left,
        text width=7.5cm, inner sep=6pt
      },
      smallpill/.style={
        draw, rounded corners=8pt, fill=white, align=center,text width=5cm, inner sep=4pt
      },
      cloud/.style={
        draw, rounded corners=12pt, fill=white, align=center,
        inner sep=6pt
      }
    }

    \node[cloud, very thick,rotate=90] (title) at (-10,0) {\textbf{Two worlds of benchmarking}};

    \node[nobox] (acad) at (-8,1.5) {\textbf{Academia}};
    \node[nobox] (ind)  at (-8,-1.5) {\textbf{Industry}};
    \node[nobox] (interested) at (-8,0) {interested in};
    \draw[dotted] (-6,0) -- node[pos=0.98, above, yshift=6pt]{\rotatebox{90}{\huge\strut academia}} node[pos=0.98, below, yshift=-6pt]{\rotatebox{90}{\huge\strut industry}} (17.5,0);

    \node[labelbox] (acad_interst) at (-2.5,1) {\textbf{Comparing algorithmic performance}};
    \node[widerlabelbox] (acad_aim) at (-2.5,3) {{\small$\bullet$} understanding\\{\small$\bullet$} progress in science};

    \node[labelbox] (ind_interst) at (-2.5,-1) {\textbf{Selecting one algorithm with minimum resources}};
    \node[widerlabelbox] (ind_aim) at (-2.5,-3) {{\small$\bullet$} solution\\{\small$\bullet$} explainability/interpretability};

    \node[widerlabelbox] (n1) at (5,1) {\textbf{Extensive set of cheap RWI benchmarks}};
    \node[widerlabelbox] (n2) at (5,-1) {\textbf{Automatic selector based on an offline database}};

    \node[smallpill] (start) at (5,-3) {Starting point for RW (benchmarking) validation};
    \node[narrowerlabelbox] (pdata) at (11,1) {\textbf{Performance data}};
    \node[narrowerlabelbox] (pdatarw) at (11,-3) {\textbf{Performance data}};

    \node[noboxwider] (triplet)  at (15,1.5) {\emph{Triplet}:\\{\small$\bullet$} problem\\{\small$\bullet$} RW features\\{\small$\bullet$} algorithm performance};
    \node[cloud] (own) at (14.4,-1.25) {Unclear ownership};
    \node[noboxwider] (pair)  at (15,-3) {\emph{\sout{Triplet} Pair}:\\{\small$\bullet$} \sout{problem}\\{\small$\bullet$} RW features\\{\small$\bullet$} algorithm performance};
    
    \draw[->, thick] (acad.south) to (interested.north);
    \draw[->, thick] (ind.north) to (interested.south);
    \draw[->, thick] (acad_interst.north) -- (acad_aim.south)
    node[midway,right] {aim};
    \draw[->, thick] (ind_interst.south) -- (ind_aim.north) node[midway,right] {aim};

    \coordinate (fork) at ($ (interested.east) + (0.5,0) $);
    \draw[thick] (interested.east) -- (fork);
    \draw[->, thick] (fork) to[out=90, in=180, looseness=1.05] (acad_interst.west);
    \draw[->, thick] (fork) to[out=-90, in=180, looseness=1.05] (ind_interst.west);
    
    \draw[->, thick] (acad_interst.east) -- (n1.west) node[midway,above] {needs};
    \draw[->, thick] (ind_interst.east) -- (n2.west) node[midway,above] {needs};
    \draw[->, thick] (n1.east) -- (pdata.west);

    \draw[->, thick] (pdata.south) to[out=-90, in=0] node[pos=0.1, right] {feeds} (n2.east);
  
    \draw[->, thick] (n2.south) -- (start.north);
    \draw[->, thick] (start.east) -- (pdatarw.west);
    
    \draw[->, thick, dashed]
      (pdatarw.north)
      to[out=90,in=-30]
      node[pos=0.1, right]{fine-tunes}
      (n1.south east);
  \end{tikzpicture}}
  \caption{Logic flowchart illustrating two worlds of benchmarking: Academia focuses on understanding and comparing algorithmic performance via RWI benchmarks, while industry aims to select effective algorithms with minimal resources using offline databases. Dashed arrows denote feedback loops where performance data fine-tune and validate benchmarks.} \label{fig:rw-bench-diagram}
\end{figure}
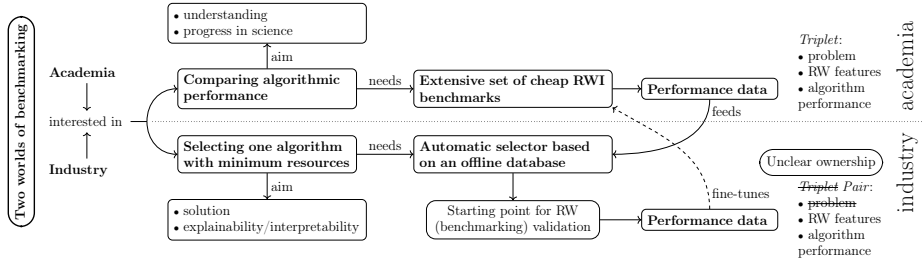

\paragraph{Different objectives and needs in Academia and Industry.} An honest assessment of the situation concerning the benchmarking objectives relating to real-world problems reveals that academia and industry typically differ strongly (see Figure~\ref{fig:rw-bench-diagram}):
From an academic perspective, one is interested in comparing algorithm performance, with the fundamental aim of understanding (e.g. which algorithms perform well on which type of problem instances?), explaining (why is that the case?) and making progress in science. 
From an industrial perspective, one is interested in selecting the single algorithm that would yield the best possible solution to the problem instance at hand while requiring minimum resources (e.g. in terms of compute time) with the aim of solving the problem and understanding the obtained solution (not the algorithm).

This results in different needs for both groups, namely an extensive set of RWI benchmark function instances for academia that are easy to evaluate and an automated optimization algorithm selector for industrial users. 
This algorithm selector would have to be based on high-level features $\bar{\mathbf{v}} = (\bar{v}_1,\ldots,\bar{v}_k)$ of real-world problems that are easily available (i.e. without any computational effort in terms of function evaluations, but rather deduced from the problem de\-fi\-ni\-tion) and can be determined by the application domain expert, i.e. the industrial end user.
Potential examples of such features have already been proposed in a real-world problem property questionnaire \cite{DBLP:conf/gecco/BlomDTMNOVN20,DBLP:series/ncs/BlomDVMNNOT23} and the rule-based algorithm selector in Nevergrad, NGOpt \cite{10.1007/978-3-031-14714-2_2}. 
Examples of features include, e.g. the number and type of variables, number and type of constraints, single vs.~multi-objective problem, the degree of parallelism for function evaluations, the total available evaluation budget, noise level and type (e.g. constant, proportional, heteroscedastic) and high-level characteristics of the problem complexity such as multimodality and (non-)separability. 
Those features need to be carefully selected and should be driven exclusively by considerations such as (i) practical requirements of the real-world application (e.g. computational effort per objective function evaluation, available wall clock time for obtaining a solution, number of parallel function evaluations possible) and (ii) features that can be determined or estimated without requiring function evaluations.

\paragraph{Towards an accessible framework for transversal benchmarking.} Assuming that the academic world has access to the above mentioned, extensive set of RWI, cheap to evaluate benchmark functions ${\cal F} = \{f_1,\ldots,f_p\}$ and a set of optimization algorithms ${\cal A} = \{A_1,\ldots, A_q\}$, it would be possible to provide the anytime performance data that captures a sufficiently large number of runs of each algorithm against each function.
The automated algorithm selector could then be based on a distance measure $d$ on feature vectors between the RWI functions, which would all be manually characterized by a feature vector $\mathbf{v}_i = (v_{i1},\ldots,v_{ik})$ ($i \in [1..p]$) as outlined above, and the real-world features $\bar{\mathbf{v}} = (\bar{v}_1,\ldots,\bar{v}_k)$ assessed by the application domain expert for a given problem at hand. 
Based on a closest match 
$i^* =\arg \min_i d(\mathbf{v}_i, \bar{\mathbf{v}})$ one would find the corresponding benchmark function $f_{i^*}$ and the best corresponding algorithm, given the available benchmarking data, number of function evaluations and level of parallelism.
The most critical components of this approach include (i) a careful selection of the high-level features, spanning across different applied fields, inspired by interaction with and experience in projects with end users, (ii) a continuously growing and carefully selected set of RWI problems from a large diversity of disciplines and (iii) the availability of proxy problems for simulation-based real-world problems. 
These three critical topics are discussed in more detail below.
\begin{description}
    \item[Selection of high-level features] In practice, collaboration between re\-searchers in Evolutionary Computation and domain experts is consultative and often needed to define the true optimization problem. 
    End users typically want quick solutions to turn into actionable results rather than exploratory studies. The information about the optimization problem they provide is usually limited to a small set of high-level features, as outlined above. A key open question remains: do similarities in these high-level features reliably translate into comparable algorithm performance or stable rankings of algorithms across problems classified as similar?
    \item[Selection of RWI optimization problems] 
    RWI problems can be grouped into synthetic, mathematically defined problems, which are easy to implement and cheap to evaluate, and simulation-based problems, which require external simulators and substantial computational resources and setup effort.
    While many examples of the former exist in the literature (e.g. 57 problems in \cite{KUMAR2020100693},  multi-objective problems in chapter 9 of \cite{Deb2001}), they often have fixed dimensionality and strong constraints, limiting their suitability for broader academic studies. 
    Truly simulation-based benchmarks remain scarce for academic benchmarking due to unfamiliarity with the toolchain (geometric representation, meshing, simulation and pre- and post-processing of results) and limited access to commercial software.
    \item[Availability of ``proxies'' for RWI problems]
    To compensate for the lack of simulation-based optimization benchmarks, researchers in academia have often relied on surrogate models that aim at representing the characteristics of the corresponding real-world problem and are cheap to evaluate (e.g. MOPTA08~\cite{jones2008large}). 
    Pipeline approaches can automate the design of such proxy functions, as proposed for structural mechanics tasks in automotive engineering~\cite{10.1145/3646554}, but typically require large initial simulator-based function evaluations and, in the end, may still fail to accurately capture the intrinsic characteristics of the real objective functions. 
    As a result, the proxy may distort key landscape features, like discontinuities and constraint-induced sharp transitions, and may lead to significant differences in algorithm performance, which is precisely what should be avoided when constructing faithful, RWI benchmarks.
\end{description}

\paragraph{Role of tooling in enabling the vision.}
Achieving this vision requires benchmarking toolboxes that support not only experimentation but also the construction and maintenance of a real-world inspired benchmark ecosystem. A first step is improving the collection and accessibility of RWI problems. While initial efforts such as the Optimization Problem Library\footnote{\url{https://openoptimizationorg.github.io/OPL/}} exist, they remain fragmented and must mature into curated, discoverable repositories that integrate smoothly with existing benchmarking workflows.

A second requirement is the systematic collection of annotated performance data on these benchmarks. Such a database should be community-maintained, interoperable with current frameworks and include basic validation and curation to ensure trustworthiness. As algorithms evolve, datasets must remain linked to the exact code versions used to generate them, enabling periodic re-validation and preventing outdated results from silently persisting, an idea already present in systems like Nevergrad's NgOpt mechanism~\cite{10.1007/978-3-031-14714-2_2}.

Finally, tooling must better support research-question–driven interpretation. Lessons from BBOB~\cite{hansen2021coco} and its documented misuse show that tools should guide users toward appropriate comparisons. For example, grouping Sphere and Ellipsoid results when studying ill-conditioning. Rather than monolithic end-to-end frameworks, a modular toolbox with clearly defined interfaces would allow researchers to assemble pipelines suited to their goals while reducing misuse and improving transparency.
These improvements are essential to make RWI benchmarks practically usable, scientifically credible and continuously aligned with the evolving needs of both academia and industry.

We realize that this vision will not be without challenges: From identifying meaningful and feasible features (in close dialogue with industrial users) to managing computational resources and coordinating efforts across distributed academic groups. Yet, these challenges also represent an exciting opportunity: By joining forces, the community can establish a living, collaborative benchmark ecosystem that continuously evolves with real-world insights and drives progress in optimization research. 

\paragraph{A forward-looking perspective.} Bringing all elements together, the proposed vision for RWI benchmarking is based on a simple but justified assumption: industrial users want to know which algorithm to run on their optimization problem, without running any time-consuming investigation beforehand. 
The information industry can realistically provide about the optimization problem, captured as high-level problem features, is based on high-level knowledge about the problem and the available resources for solving it. 
With these features as input, academia can take on the complementary responsibility: curating a diverse collection of RWI problems, characterizing each by its feature vector, and generating a common database of anytime performance data by running a broad set of algorithms across these problems. Such a database would form a foundation for algorithm selection. By identifying the benchmark instance whose feature vector most closely matches that of a user’s problem, one can recommend the algorithm that performs best under comparable conditions (e.g. evaluation budget, degree of parallelism).
Importantly, as illustrated in Figure \ref{fig:rw-bench-diagram}, this process is not one-directional. The dashed arrows represent a feedback mechanism whereby performance data collected via the tooling discussed above cab feed back into academia to adjust and fine-tune the taxonomy and construction of cheap RWI benchmarks, ensuring that benchmark generators and feature spaces remain aligned with emerging real-world problem characteristics. In other words, real-world performance data serves as a reality check, refining benchmark design so that academic testbeds evolve in tandem with industrial needs rather than diverging from them.

\section{Conclusions and Outlook}
Benchmarking has shaped empirical research in Evolutionary Computation for decades, yet its impact on real-world optimization remains limited. Synthetic benchmarks such as BBOB and the IOH suite have been invaluable for understanding algorithmic behaviour, but they do not adequately represent the diversity, constraints and information limitations of continuous and mixed-integer optimization problems arising in practice. As a result, academic benchmarking has drifted away from the needs of industrial users, who must make high-stakes decisions under tight budgets and incomplete problem knowledge.

This paper argued that progress requires a shift toward principled, real-world-inspired benchmarking. Such a shift hinges on three elements: 1) a taxo\-nomy of high-level problem features that practitioners can specify without expensive evaluations, 2) curated collections of RWI benchmark problems spanning diverse application domains, 3) community-driven tooling and data repositories that support trustworthy experimentation and enable informed solver selection. We believe these components are essential for narrowing the gap between academic insights and industrial relevance.

Moving forward, the main challenge is coordination rather than methodology. No single group can design, collect and maintain the necessary benchmark ecosystem. Instead, sustained community effort is needed, through shared datasets, collaborative benchmark design and regular re-validation of results, to establish a living benchmark infrastructure that evolves with emerging applications. Industrial performance data, in turn, should feed back into academia to refine taxonomies, proxy generators and benchmark families over time.

We hope this paper contributes to a more impact-oriented benchmarking culture. If benchmarking is to matter, for research, for industry and for the credibility of our field, it must become a shared, long-term endeavour grounded in realistic problems, transparent tooling, and clear intent.

\subsubsection*{Disclosure of Interests.}
The authors have no competing interests to declare that are relevant to the content of this article.

\vspace{10pt}

\begin{credits}
This article is the outomce of \href{https://www.dagstuhl.de/en/seminars/seminar-calendar/seminar-details/25444}{Dagstuhl Research Meeting 25444 ``Better Benchmarking Setups for optimization: Design, Curation and Long-Term Evolution''}. All the authors thank Schloss Dagstuhl for hosting this event.

This article is based upon work from COST Action CA22137 ROAR-NET, supported by COST  (European Cooperation in Science and Technology). 

%
\end{credits}
%
%
%
\bibliographystyle{splncs04}
\bibliography{bib.bib}

@article{hansen2021coco,
  title={COCO: A platform for comparing continuous optimizers in a black-box setting},
  author={Hansen, Nikolaus and Auger, Anne and Ros, Raymond and Mersmann, Olaf and Tu{\v{s}}ar, Tea and Brockhoff, Dimo},
  journal={Optimization Methods and Software},
  volume={36},
  number={1},
  pages={114--144},
  year={2021},
  publisher={Taylor \& Francis}
}

@article{back2023evolutionary,
  title={Evolutionary algorithms for parameter optimization—thirty years later},
  author={B{\"a}ck, Thomas HW and Kononova, Anna V and van Stein, Bas and Wang, Hao and Antonov, Kirill A and Kalkreuth, Roman T and de Nobel, Jacob and Vermetten, Diederick and de Winter, Roy and Ye, Furong},
  journal={Evolutionary Computation},
  volume={31},
  number={2},
  pages={81--122},
  year={2023},
  publisher={MIT Press}
}

@inproceedings{vskvorc2019cec,
  title={CEC real-parameter optimization competitions: Progress from 2013 to 2018},
  author={{\v{S}}kvorc, Urban and Eftimov, Tome and Koro{\v{s}}ec, Peter},
  booktitle={2019 IEEE congress on evolutionary computation (CEC)},
  pages={3126--3133},
  year={2019},
  organization={IEEE}
}

@inproceedings{lacroix2019limitations,
  title={Limitations of benchmark sets and landscape features for algorithm selection and performance prediction},
  author={Lacroix, Benjamin and McCall, John},
  booktitle={Proceedings of the Genetic and Evolutionary Computation Conference Companion},
  pages={261--262},
  year={2019}
}

@article{volz2023tools,
  title={Tools for landscape analysis of optimisation problems in procedural content generation for games},
  author={Volz, Vanessa and Naujoks, Boris and Kerschke, Pascal and Tu{\v{s}}ar, Tea},
  journal={Applied Soft Computing},
  volume={136},
  pages={110121},
  year={2023},
  publisher={Elsevier}
}

@inproceedings{kerschke2015detecting,
  title={Detecting funnel structures by means of exploratory landscape analysis},
  author={Kerschke, Pascal and Preuss, Mike and Wessing, Simon and Trautmann, Heike},
  booktitle={Proceedings of the 2015 Annual Conference on Genetic and Evolutionary Computation},
  pages={265--272},
  year={2015}
}

@article{shehadeh2025benchmarking,
  title={Benchmarking global optimization techniques for unmanned aerial vehicle path planning},
  author={Shehadeh, Mhd Ali and Kudela, Jakub},
  journal={Expert Systems with Applications},
  pages={128645},
  year={2025},
  publisher={Elsevier}
}

@inproceedings{kuudela2024benchmarking,
  title={Benchmarking derivative-free global optimization methods on variable dimension robotics problems},
  author={Kudela, Jakub and Ju{\v{r}}{\'\i}{\v{c}}ek, Martin and Par{\'a}k, Roman and Tzanetos, Alexandros and Matou{\v{s}}ek, Radomil},
  booktitle={2024 IEEE Congress on Evolutionary Computation (CEC)},
  pages={1--8},
  year={2024},
  publisher={IEEE}
}

@inproceedings{DBLP:conf/gecco/MersmannBTPWR11,
  author       = {Olaf Mersmann and
                  Bernd Bischl and
                  Heike Trautmann and
                  Mike Preuss and
                  Claus Weihs and
                  G{\"{u}}nter Rudolph},
  editor       = {Natalio Krasnogor and
                  Pier Luca Lanzi},
  title        = {Exploratory landscape analysis},
  booktitle    = {13th Annual Genetic and Evolutionary Computation Conference, {GECCO}
                  2011, Proceedings, Dublin, Ireland, July 12-16, 2011},
  pages        = {829--836},
  publisher    = {{ACM}},
  year         = {2011},
  url          = {https://doi.org/10.1145/2001576.2001690},
  doi          = {10.1145/2001576.2001690},
  bibsource    = {dblp computer science bibliography, https://dblp.org}
}

@article{DBLP:journals/swevo/PetelinCE24,
  author       = {Gasper Petelin and
                  Gjorgjina Cenikj and
                  Tome Eftimov},
  title        = {TinyTLA: Topological landscape analysis for optimization problem classification
                  in a limited sample setting},
  journal      = {Swarm Evol. Comput.},
  volume       = {84},
  pages        = {101448},
  year         = {2024},
  url          = {https://doi.org/10.1016/j.swevo.2023.101448},
  doi          = {10.1016/J.SWEVO.2023.101448},
  bibsource    = {dblp computer science bibliography, https://dblp.org}
}

@article{DBLP:journals/corr/abs-2401-01192,
  author       = {Moritz Vinzent Seiler and
                  Pascal Kerschke and
                  Heike Trautmann},
  title        = {Deep-ELA: Deep Exploratory Landscape Analysis with Self-Supervised
                  Pretrained Transformers for Single- and Multi-Objective Continuous
                  Optimization Problems},
  journal      = {CoRR},
  volume       = {abs/2401.01192},
  year         = {2024},
  url          = {https://doi.org/10.48550/arXiv.2401.01192},
  doi          = {10.48550/ARXIV.2401.01192},
  eprinttype    = {arXiv},
  eprint       = {2401.01192},
  bibsource    = {dblp computer science bibliography, https://dblp.org}
}

@misc{CEC2017,
author={Wu, Guohua and Mallipeddi, Rammohan and Suganthan, Ponnuthurai N.},
title={{Problem Definitions and Evaluation Criteria for the {CEC} 2017 Competition on Constrained Single Objective Real-Parameter Optimization}},
note={Technical Report, Nanyang Technological University, Singapore},
url={https://github.com/P-N-Suganthan/CEC2017},
year={2017}
}

@inproceedings{DBLP:conf/foga/DerbelLVAT19,
  author       = {Bilel Derbel and
                  Arnaud Liefooghe and
                  S{\'{e}}bastien V{\'{e}}rel and
                  Hern{\'{a}}n E. Aguirre and
                  Kiyoshi Tanaka},
  editor       = {Tobias Friedrich and
                  Carola Doerr and
                  Dirk V. Arnold},
  title        = {New features for continuous exploratory landscape analysis based on
                  the {SOO} tree},
  booktitle    = {Proceedings of the 15th {ACM/SIGEVO} Conference on Foundations of
                  Genetic Algorithms, {FOGA} 2019, Potsdam, Germany, August 27-29, 2019},
  pages        = {72--86},
  publisher    = {{ACM}},
  year         = {2019},
  url          = {https://doi.org/10.1145/3299904.3340308},
  doi          = {10.1145/3299904.3340308},
  bibsource    = {dblp computer science bibliography, https://dblp.org}
}

@inproceedings{DBLP:conf/evoW/RenauDDD21,
  author       = {Quentin Renau and
                  Johann Dr{\'{e}}o and
                  Carola Doerr and
                  Benjamin Doerr},
  editor       = {Pedro A. Castillo and
                  Juan Luis Jim{\'{e}}nez Laredo},
  title        = {Towards Explainable Exploratory Landscape Analysis: Extreme Feature
                  Selection for Classifying {BBOB} Functions},
  booktitle    = {Applications of Evolutionary Computation - 24th International Conference,
                  EvoApplications 2021, Held as Part of EvoStar 2021, Virtual Event,
                  April 7-9, 2021, Proceedings},
  series       = {Lecture Notes in Computer Science},
  volume       = {12694},
  pages        = {17--33},
  publisher    = {Springer},
  year         = {2021},
  url          = {https://doi.org/10.1007/978-3-030-72699-7\_2},
  doi          = {10.1007/978-3-030-72699-7\_2},
  bibsource    = {dblp computer science bibliography, https://dblp.org}
}

@article{DBLP:journals/swevo/CenikjPSCE25,
  author       = {Gjorgjina Cenikj and
                  Gasper Petelin and
                  Moritz Seiler and
                  Nikola Cenikj and
                  Tome Eftimov},
  title        = {Landscape features in single-objective continuous optimization: Have
                  we hit a wall in algorithm selection generalization?},
  journal      = {Swarm Evol. Comput.},
  volume       = {94},
  pages        = {101894},
  year         = {2025},
  url          = {https://doi.org/10.1016/j.swevo.2025.101894},
  doi          = {10.1016/J.SWEVO.2025.101894},
  bibsource    = {dblp computer science bibliography, https://dblp.org}
}

@incollection{DBLP:series/ncs/BlomDVMNNOT23,
  author       = {Koen van der Blom and
                  Timo M. Deist and
                  Vanessa Volz and
                  Mariapia Marchi and
                  Yusuke Nojima and
                  Boris Naujoks and
                  Akira Oyama and
                  Tea Tusar},
  editor       = {Dimo Brockhoff and
                  Michael Emmerich and
                  Boris Naujoks and
                  Robin C. Purshouse},
  title        = {Identifying Properties of Real-World Optimisation Problems Through
                  a Questionnaire},
  booktitle    = {Many-Criteria Optimization and Decision Analysis: State-of-the-Art,
                  Present Challenges, and Future Perspectives},
  series       = {Natural Computing Series},
  pages        = {59--80},
  publisher    = {Springer},
  year         = {2023},
  url          = {https://doi.org/10.1007/978-3-031-25263-1\_3},
  doi          = {10.1007/978-3-031-25263-1\_3},
  bibsource    = {dblp computer science bibliography, https://dblp.org}
}

@inproceedings{10.1007/978-3-031-14714-2_2,
author = {Trajanov, Risto and Nikolikj, Ana and Cenikj, Gjorgjina and Teytaud, Fabien and Videau, Mathurin and Teytaud, Olivier and Eftimov, Tome and L\'{o}pez-Ib\'{a}\~{n}ez, Manuel and Doerr, Carola},
title = {Improving Nevergrad’s Algorithm Selection Wizard NGOpt Through Automated Algorithm Configuration},
year = {2022},
isbn = {978-3-031-14713-5},
publisher = {Springer-Verlag},
address = {Berlin, Heidelberg},
url = {https://doi.org/10.1007/978-3-031-14714-2_2},
doi = {10.1007/978-3-031-14714-2_2},
booktitle = {Parallel Problem Solving from Nature – PPSN XVII: 17th International Conference, PPSN 2022, Dortmund, Germany, September 10–14, 2022, Proceedings, Part I},
pages = {18–31},
numpages = {14},
location = {Dortmund, Germany}
}

@inproceedings{DBLP:conf/gecco/BlomDTMNOVN20,
  author       = {Koen van der Blom and
                  Timo M. Deist and
                  Tea Tusar and
                  Mariapia Marchi and
                  Yusuke Nojima and
                  Akira Oyama and
                  Vanessa Volz and
                  Boris Naujoks},
  editor       = {Carlos Artemio Coello Coello},
  title        = {Towards realistic optimization benchmarks: a questionnaire on the
                  properties of real-world problems},
  booktitle    = {{GECCO} '20: Genetic and Evolutionary Computation Conference, Companion
                  Volume, Canc{\'{u}}n, Mexico, July 8-12, 2020},
  pages        = {293--294},
  publisher    = {{ACM}},
  year         = {2020},
  url          = {https://doi.org/10.1145/3377929.3389974},
  doi          = {10.1145/3377929.3389974},
  bibsource    = {dblp computer science bibliography, https://dblp.org}
}

@techreport{bbob2019,
    author = {Finck, Steffen and Hansen, Nikolaus and Ros, Raymond and Auger, Anne},
    title = {Real-Parameter Black-Box Optimization Benchmarking 2009: Noiseless Functions Definitions},
    institution = {INRIA},
    year = {2009},
    number = {RR-6829},
    note = {Updated version as of February 2019},
    url = {https://inria.hal.science/inria-00362633v2/document}
}

@article{10.1145/3646554,
author = {Long, Fu Xing and van Stein, Bas and Frenzel, Moritz and Krause, Peter and Gitterle, Markus and B\"{a}ck, Thomas},
title = {Generating Cheap Representative Functions for Expensive Automotive Crashworthiness Optimization},
year = {2024},
issue_date = {June 2024},
publisher = {Association for Computing Machinery},
address = {New York, NY, USA},
volume = {4},
number = {2},
url = {https://doi.org/10.1145/3646554},
doi = {10.1145/3646554},
journal = {ACM Trans. Evol. Learn. Optim.},
month = jun,
articleno = {9},
numpages = {26}
}

@article{KUMAR2020100693,
title = {A test-suite of non-convex constrained optimization problems from the real-world and some baseline results},
journal = {Swarm and Evolutionary Computation},
volume = {56},
pages = {100693},
year = {2020},
issn = {2210-6502},
doi = {https://doi.org/10.1016/j.swevo.2020.100693},
url = {https://www.sciencedirect.com/science/article/pii/S2210650219308946},
author = {Abhishek Kumar and Guohua Wu and Mostafa Z. Ali and Rammohan Mallipeddi and Ponnuthurai Nagaratnam Suganthan and Swagatam Das}
}

@book{Deb2001,
author = {Deb, Kalyan},
year = {2001},
month = {01},
publisher = {Wiley},
pages = {},
title = {Multiobjective Optimization Using Evolutionary Algorithms}
}

@inproceedings{jones2008large,
  title={Large-scale multi-disciplinary mass optimization in the auto industry},
  author={Jones, Donald R},
  booktitle={MOPTA 2008 Conference (20 August 2008)},
  volume={64},
  year={2008}
}

@misc{chas10a,
	author = {N. Chase and M. Rademacher and  E. Goodman and R. Averill and R. Sidhu},
	title = {A Benchmark Study of Optimization Search Algorithms},
	year = {2010}}

@misc{bart20gArxiv,
	author = {Thomas Bartz-Beielstein and Carola Doerr and Jakob Bossek and Sowmya Chandrasekaran and Tome Eftimov and Andreas Fischbach and Pascal Kerschke and Manuel Lopez-Ibanez and Katherine M. Malan and Jason H. Moore and Boris Naujoks and Patryk Orzechowski and Vanessa Volz and Markus Wagner and Thomas Weise},
	howpublished = {arXiv},
	month = {07},
	title = {Benchmarking in Optimization: Best Practice and Open Issues},
	year = {2020}}

@article{grimme2021peeking,
  title={Peeking beyond peaks: Challenges and research potentials of continuous multimodal multi-objective optimization},
  author={Grimme, Christian and Kerschke, Pascal and Aspar, Pelin and Trautmann, Heike and Preuss, Mike and Deutz, Andre H and Wang, Hao and Emmerich, Michael},
  journal={Computers \& Operations Research},
  volume={136},
  pages={105489},
  year={2021},
  publisher={Elsevier}
}

@inproceedings{glasmachers2019challenges,
  title={Challenges of convex quadratic bi-objective benchmark problems},
  author={Glasmachers, Tobias},
  booktitle={Proceedings of the Genetic and Evolutionary Computation Conference},
  pages={559--567},
  year={2019}
}

@article{ABC+21,
  author = {Audet, C. and Bigeon, J. and Cartier, D. and Le Digabel, S. and Salomon, L.},
  title = {Performance indicators in multiobjective optimization},
  pages = {397-422},
  journal = {European Journal of Operational Research},
  volume = 292,
  number = 2,
  year = 2021,
  doi = {10.1016/j.ejor.2020.11.016}
}

@article{LY19,
author={Li, M. and Yao, X.},
title={Quality Evaluation of Solution Sets in Multiobjective Optimisation: A Survey}, 
journal={ACM Computing Surveys}, 
volume=52,
number=2,
year=2019,
pages={1--38}, 
note={article 26},
doi={10.1145/3300148}
}

@ARTICLE{ShaEtAl21,
  author={Shang, Ke and Ishibuchi, Hisao and He, Linjun and Pang, Lie Meng},
  journal={IEEE Transactions on Evolutionary Computation}, 
  title={A Survey on the Hypervolume Indicator in Evolutionary Multiobjective Optimization}, 
  year={2021},
  volume={25},
  number={1},
  pages={1-20},
  doi={10.1109/TEVC.2020.3013290}}

@inproceedings{Renau2024,
author = {Renau, Quentin and Dreo, Johann and Hart, Emma},
title = {Ealain: A Camera Simulation Tool to Generate Instances for Multiple Classes of Optimisation Problem},
year = {2024},
isbn = {9798400704956},
publisher = {Association for Computing Machinery},
address = {New York, NY, USA},
url = {https://doi.org/10.1145/3638530.3654299},
doi = {10.1145/3638530.3654299},
booktitle = {Proceedings of the Genetic and Evolutionary Computation Conference Companion},
pages = {151–154},
numpages = {4},
location = {Melbourne, VIC, Australia},
series = {GECCO '24 Companion}
}

@inproceedings{Thomaser2023,
  author       = {Andr{\'{e}} Thomaser and
                  Marc{-}Eric Vogt and
                  Thomas B{\"{a}}ck and
                  Anna V. Kononova},
  editor       = {Niki van Stein and
                  Francesco Marcelloni and
                  H. K. Lam and
                  Marie Cottrell and
                  Joaquim Filipe},
  title        = {Real-World Optimization Benchmark from Vehicle Dynamics: Specification
                  of Problems in 2D and Methodology for Transferring (Meta-)Optimized
                  Algorithm Parameters},
  booktitle    = {Proceedings of the 15th International Joint Conference on Computational
                  Intelligence, {IJCCI} 2023, Rome, Italy, November 13-15, 2023},
  pages        = {31--40},
  publisher    = {{SCITEPRESS}},
  year         = {2023},
  url          = {https://doi.org/10.5220/0012158000003595},
  doi          = {10.5220/0012158000003595},
  bibsource    = {dblp computer science bibliography, https://dblp.org}
}

@article{Tanabe2020,
title = {An easy-to-use real-world multi-objective optimization problem suite},
journal = {Applied Soft Computing},
volume = {89},
pages = {106078},
year = {2020},
issn = {1568-4946},
doi = {https://doi.org/10.1016/j.asoc.2020.106078},
url = {https://www.sciencedirect.com/science/article/pii/S1568494620300181},
author = {Ryoji Tanabe and Hisao Ishibuchi},
}

@article{He2020,
  author    = {He, C. and Tian, Y. and Wang, H. and Zhang, X. and Zhang, Y.},
  title     = {A repository of real-world datasets for data-driven evolutionary multiobjective optimization},
  journal   = {Complex \& Intelligent Systems},
  year      = {2020},
  volume    = {6},
  pages     = {189--197},
  doi       = {10.1007/s40747-019-00126-2},
  url       = {https://doi.org/10.1007/s40747-019-00126-2}
}

@inproceedings{Volz2019,
author = {Volz, Vanessa and Naujoks, Boris and Kerschke, Pascal and Tu\v{s}ar, Tea},
title = {Single- and multi-objective game-benchmark for evolutionary algorithms},
year = {2019},
isbn = {9781450361118},
publisher = {Association for Computing Machinery},
address = {New York, NY, USA},
url = {https://doi.org/10.1145/3321707.3321805},
doi = {10.1145/3321707.3321805},
booktitle = {Proceedings of the Genetic and Evolutionary Computation Conference},
pages = {647–655},
numpages = {9},
location = {Prague, Czech Republic},
series = {GECCO '19}
}

@ARTICLE{Picard2021,
  author={Picard, Cyril and Schiffmann, Jürg},
  journal={IEEE Transactions on Evolutionary Computation}, 
  title={Realistic Constrained Multiobjective Optimization Benchmark Problems From Design}, 
  year={2021},
  volume={25},
  number={2},
  pages={234-246},
  doi={10.1109/TEVC.2020.3020046}}

@misc{Trabucco2022,
      title={Design-Bench: Benchmarks for Data-Driven Offline Model-Based Optimization}, 
      author={Brandon Trabucco and Xinyang Geng and Aviral Kumar and Sergey Levine},
      year={2022},
      eprint={2202.08450},
      archivePrefix={arXiv},
      primaryClass={cs.LG},
      url={https://arxiv.org/abs/2202.08450}, 
}

@inproceedings{
Qian2025,
title={{SOO}-Bench: Benchmarks for Evaluating the Stability of Offline Black-Box Optimization},
author={Hong Qian and Yiyi Zhu and Xiang Shu and Shuo Liu and Yaolin Wen and Xin An and Huakang Lu and Aimin Zhou and Ke Tang and Yang Yu},
booktitle={The Thirteenth International Conference on Learning Representations},
year={2025},
url={https://openreview.net/forum?id=bqf0aCF3Dd}
}

@article{rardin2001experimental,
  title={Experimental evaluation of heuristic optimization algorithms: A tutorial},
  author={Rardin, Ronald L and Uzsoy, Reha},
  journal={Journal of Heuristics},
  volume={7},
  pages={261--304},
  year={2001},
  publisher={Springer}
}

@article{whitley1996evaluating,
  title={Evaluating evolutionary algorithms},
  author={Whitley, Darrell and Rana, Soraya and Dzubera, John and Mathias, Keith E},
  journal={Artificial intelligence},
  volume={85},
  number={1-2},
  pages={245--276},
  year={1996},
  publisher={Elsevier}
}

@article{johnson2002experimental,
  title={A theoretician's guide to the experimental analysis of algorithms},
  author={Johnson, David S},
  journal={Data Structures, Near Neighbor Searches, and Methodology: 5th and 6th DIMACS Implementation Challenges},
  volume={59},
  pages={215--250},
  year={2001}
}

@article{beiranvand2017best,
  title={Best practices for comparing optimization algorithms},
  author={Beiranvand, Vahid and Hare, Warren and Lucet, Yves},
  journal={Optimization and Engineering},
  volume={18},
  number={4},
  pages={815--848},
  year={2017},
  publisher={Springer}
}

@article{HELLWIG2019bench,
title = {Benchmarking evolutionary algorithms for single objective real-valued constrained optimization – A critical review},
journal = {Swarm and Evolutionary Computation},
volume = {44},
pages = {927-944},
year = {2019},
issn = {2210-6502},
doi = {https://doi.org/10.1016/j.swevo.2018.10.002},
url = {https://www.sciencedirect.com/science/article/pii/S2210650218305406},
author = {Michael Hellwig and Hans-Georg Beyer}
}

@ARTICLE{IOHprofiler,
  author = {Carola Doerr and Hao Wang and Furong Ye and Sander van Rijn and Thomas B{\"a}ck},
  title = {{IOHprofiler: A Benchmarking and Profiling Tool for Iterative Optimization Heuristics}},
  journal = {arXiv e-prints:1810.05281},
  archivePrefix = "arXiv",
  eprint = {1810.05281},
  year = 2018,
  month = oct,
  url = {https://arxiv.org/abs/1810.05281}
}

@incollection{JFS97,
author = {De Jong, Kenneth and Fogel, David and Schwefel, Hans-Paul},
year = {1997},
pages = {A2.3:1-12},
title = {A history of evolutionary computation},
booktitle = {Handbook of Evolutionary Computation, IOP Publishing Ltd},
editor={Thomas B{\"a}ck and David B. Fogel and Zbigniew Michalewicz},
publisher={Oxford University Press and the Institute of Physics, UK},
doi={https://doi.org/10.1201/9780367802486}
}

@ARTICLE{DPA+02,
  author={Deb, K. and Pratap, A. and Agarwal, S. and Meyarivan, T.},
  journal={IEEE Transactions on Evolutionary Computation}, 
  title={A fast and elitist multiobjective genetic algorithm: NSGA-II}, 
  year={2002},
  volume={6},
  number={2},
  pages={182-197},
  doi={10.1109/4235.996017}}

@article{BNE07,
title = {SMS-EMOA: Multiobjective selection based on dominated hypervolume},
journal = {European Journal of Operational Research},
volume = {181},
number = {3},
pages = {1653-1669},
year = {2007},
issn = {0377-2217},
doi = {https://doi.org/10.1016/j.ejor.2006.08.008},
url = {https://www.sciencedirect.com/science/article/pii/S0377221706005443},
author = {Nicola Beume and Boris Naujoks and Michael Emmerich}
}

@article{NFLT,
  author        = "D. Wolpert and W. Macready",
  title         = "No free lunch theorems for optimization",
  journal       = "IEEE Transactions on Evolutionary Computation",
  volume        = "1",
  issue         = "1",
  year          = "1997",
  doi ="10.1109/4235.585893",
  pages         = "67--82"
}

@article{Rowe2009,
    author = {Rowe, Jon E. and Vose, M. D. and Wright, Alden H.},
    title = {Reinterpreting No Free Lunch},
    journal = {Evolutionary Computation},
    volume = {17},
    number = {1},
    pages = {117-129},
    year = {2009},
    month = {03},
    issn = {1063-6560},
    doi = {10.1162/evco.2009.17.1.117},
    url = {https://doi.org/10.1162/evco.2009.17.1.117},
    eprint = {https://direct.mit.edu/evco/article-pdf/17/1/117/1493818/evco.2009.17.1.117.pdf},
}

@INPROCEEDINGS{Knowles2002_metrics,
  author={Knowles, J. and Corne, D.},
  booktitle={Proceedings of the 2002 Congress on Evolutionary Computation. CEC'02 (Cat. No.02TH8600)}, 
  title={On metrics for comparing nondominated sets}, 
  year={2002},
  volume={1},
  number={},
  pages={711-716 vol.1},
  doi={10.1109/CEC.2002.1007013}}
\end{document}